# A Hybrid Input based Deep Reinforcement Learning for Lane Change Decision-Making of Autonomous Vehicle

Gao Ziteng, Qu Jiaqi, Chen Chaoyu*

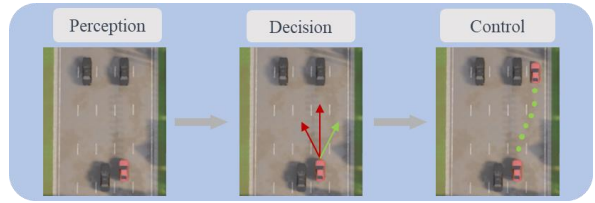

**Figure. 1:** Sequence steps of lane change

*Abstract—* Lane change decision-making for autonomous vehicles is a complex but high-reward behavior. In this paper, we propose a hybrid input based deep reinforcement learning (DRL) algorithm, which realizes abstract lane change decisions and lane change actions for autonomous vehicles within traffic flow. Firstly, a surrounding vehicles trajectory prediction method is proposed to reduce the risk of future behavior of surrounding vehicles to ego vehicle, and the prediction results are input into the reinforcement learning model as additional information. Secondly, to comprehensively leverage environmental information, the model extracts feature from high-dimensional images and low-dimensional sensor data simultaneously. The fusion of surrounding vehicle trajectory prediction and multi-modal information are used as state space of reinforcement learning to improve the rationality of lane change decision. Finally, we integrate reinforcement learning macro decisions with end-to-end vehicle control to achieve a holistic lane change process. Experiments were conducted within the CARLA simulator, and the results demonstrated that the utilization of a hybrid state space significantly enhances the safety of vehicle lane change decisions.

## I. INTRODUCTION

Autonomous vehicles (AVs) have rapidly emerged in recent years [1, 2], and researchers believe that this will have extensive impacts on the entire society. Within all processes of an autonomous vehicle, lane change stands as a foundational and essential component [3], offering substantial potential to enhance driving efficiency which gathering significant research interest. In this work, as shown in Fig. 1, we separate the lane change behavior of autonomous vehicles into three sequential steps: perception, decision-making, and control.

Compared with other simple tasks such as following other vehicles [4], the challenge of lane change decision-making primarily lies in how to make a rationale and safety decision. Decision-making tasks have historically presented elevated research complexity due to their abstract nature, and the intricate environment within traffic flow further amplifies the difficulty of making decisions. Generating safe lane change decisions at opportune moments not only enhances driving efficiency but also improves overall traffic safety. Therefore, finding an effective method to generate safe lane change decisions is of great importance.

Methods for autonomous lane change decision-making can be categorized into two groups: 1) cost-function-based methods and 2) learning-based methods. Cost-function-based methods often generate optimal decisions by minimizing the cost value, but they tend to be computationally expensive, conservative in strategy, and exhibit poor generalization [5]. Within the learning-based methods, deep reinforcement learning (DRL) emerges as a potent decision-generation technique. Given that driving behavior can be modeled as a partially observable Markov decision process (POMDP) [6], this mathematical framework of sequential decision-making forms the foundation for the implementation of DRL algorithms.

Although reinforcement learning provides favorable tools for lane change driving decision-making, the effectiveness of lane change decisions still depends on the quality of the information source. In reality, autonomous vehicles often utilize a combination of sensor and camera information to complement each other, achieving more accurate perception and providing comprehensive and effective information for subsequent decision-making tasks [7]. Therefore, an ideal vehicle control decision-making model should be able to accept various types of input data, while adapting to a mixture of high-dimensional and low-dimensional inputs presents a challenge. Currently, most deep reinforcement learning methods still only leverage low-dimensional information. In the study [8], a 37-dimensional low-dimensional state space encompassing positions, velocities, angular velocities, throttle, and steering positions of the ego vehicle and surrounding vehicles was employed. However, Kendall et al. [9] proposed that the introduction of image information can effectively enhance the performance of autonomous driving tasks.

On the other hand, the interactions among vehicles within traffic flow significantly impact the behavior of autonomous vehicles [10]. Therefore, while making lane change decisions, taking into account the future trajectories of surrounding vehicles appropriately, implementing an implicit communication vehicle interaction modeling can improve the rationality and safety of lane change decisions. In summary, a more comprehensive information acquisition strategy and the inclusion of vehicle interaction effects within the state space can effectively enhance the reasonability and safety of lane change decision-making for ego vehicle.

Gao Ziteng is with the Department of Mechanical Engineering, College of Design and Engineering, National University of Singapore (e-mail: e1010863@u.nus.edu).

Qu Jiaqi is with the School of Biomedical Engineering, Shanghai Jiao Tong University (e-mail: jqqu01@sjtu.edu.cn).

Chen Chaoyu, Peter, is an associate professor with the Department of Mechanical Engineering, College of Design and Engineering, National University of Singapore (e-mail: mpechenp@nus.edu.sg).

*Corresponding author: Chen Chaoyu, Peter

Therefore, we present a hybrid-input based deep reinforcement learning for autonomous vehicle lane change decision-making, aiming to achieve optimal lane change decisions within traffic flow. Specifically, to consider the influence of surrounding vehicles, we predict the trajectories of surrounding vehicles. Furthermore, to make comprehensive use of the collected information, our model simultaneously processes low-dimensional data from basic sensors and high-dimensional image data from cameras. a convolutional neural network (CNN) based model is employed for feature extraction and dimension reduction of the images. Moreover, by integrating the information from trajectory prediction about surrounding vehicles, high-dimensional semantic images, and current low-dimensional sensor data, a comprehensive state space is constructed for the DRL model to make the lane change decision. Finally, by combing the lane change decision and rule-based control algorithms, safe lane change behavior of the vehicle is achieved. All experiments were conducted on the CARLA simulator, and the results demonstrated that the proposed method can achieve continuous and safe lane change behavior.

In summary, we present the following key contributions:

1) A trajectory prediction of surrounding vehicles is realized by a transformer network and the predicted results are combined into the state space of reinforcement learning to implicitly model the vehicle interaction impact on ego vehicle in traffic flow.

2) A DRL model is designed to fully use the hybrid environmental observation which includes low-dimensional data from basic sensors and high-dimensional image from RGB camera.

3) The lane change decision model based on deep reinforcement learning (Hybrid-PPO) is proposed, which combined with a rule-based end-to-end controller, and these achieved the entire lane change process of ego vehicles.

## II. RELATED WORK

### A. Lane Change Decision-Making

According to the existing research, common methods for lane change decision-making tasks have two main categories: cost-function-based method and reinforcement-learning-based method.

The early lane change decision is mainly based on cost function. In general, researchers treat lane change decision as an optimization problem of sequence control. The decision is obtained by calculating the method which can minimize the cost function value under the constraint. Specific methods, such as using MPC [11], constitute a cost function through vehicle kinematics, detection of lane change risk and other constraints, and finally realize optimal lane change decision.

The main problem of most decision-making methods based on cost function is that they do not provide a comprehensive explanation for lane change decision. Therefore, most of them can only match some specific driving situations. When there is uncertainty in the environment, many different approaches are required. In addition, such models usually require high computational resource.

Most reinforcement-learning-based methods treat lane change decision task as a partially observable Markov decision process (POMDP) to solve the uncertainty of complex traffic environment [12]. Reinforcement learning agent collects environmental information through trial and error, and finally obtains the best policy for decision-making. Some deep reinforcement learning approaches directly train end-to-end control policy [13, 14], which use DQN or A2C approaches to learn decision-making strategies with optimal value functions. Some other methods focus on abstract decision-making, for example using different accelerations as discrete action space to model abstract decisions [15]. In recent years, some complex deep reinforcement learning methods such as DDPG and PPO have also begun to be used for lane change decision-making in the field of autonomous driving.

In this work, we improve the performance of the model through reasonable hybrid state space and maximize the rationality of the lane change decision.

### B. Trajectory Prediction

Predicting the future state of surrounding vehicles can effectively enhance the ego vehicle's capacity to anticipate potential hazards [16]. The anticipated behaviors of vehicles on the road are shaped by factors such as high inertia, driving rules, and road geometry, which brings the possibility of trajectory prediction.

Traditional vehicle trajectory prediction is mainly based on the physical model of the vehicle [17] and the expectation of vehicle control. With the rapid development of deep learning, trajectory prediction with consideration of the interaction among vehicles in traffic flow began to be researched. At present, vehicle trajectory prediction based on deep learning has widespread applications, common models include RNN [18] and LSTM [19]. These models treat trajectories as time series data for prediction purposes. Some works also incorporate CNN [20] to maximize the utilization of environmental information. However, the ability of these method to extract information from long sequences remains constrained. Recently, some work has focused on the research of transformer-based vehicle trajectory prediction due to the effectiveness of transformer's self-attention mechanism in preserving long-term sequence information [21].

To summary, conducting trajectory prediction of surrounding vehicles can provide information for decision-making during lane changes. Ego vehicle can avoid potential collision and then enhance the security of the lane change behavior.

## III. PROBLEM FORMULATION

### A. Markov Decision Process

Markov Decision Process (MDP) is the underlying mathematical model of reinforcement learning, which is described by a five-element tuple $M: (s, a, R, P, \gamma)$. Among them, $S$ is the set of state, $A$ is the set of action, $R(s, a)$ is the reward function, $P(s' | s, a)$ is the state transition probability,

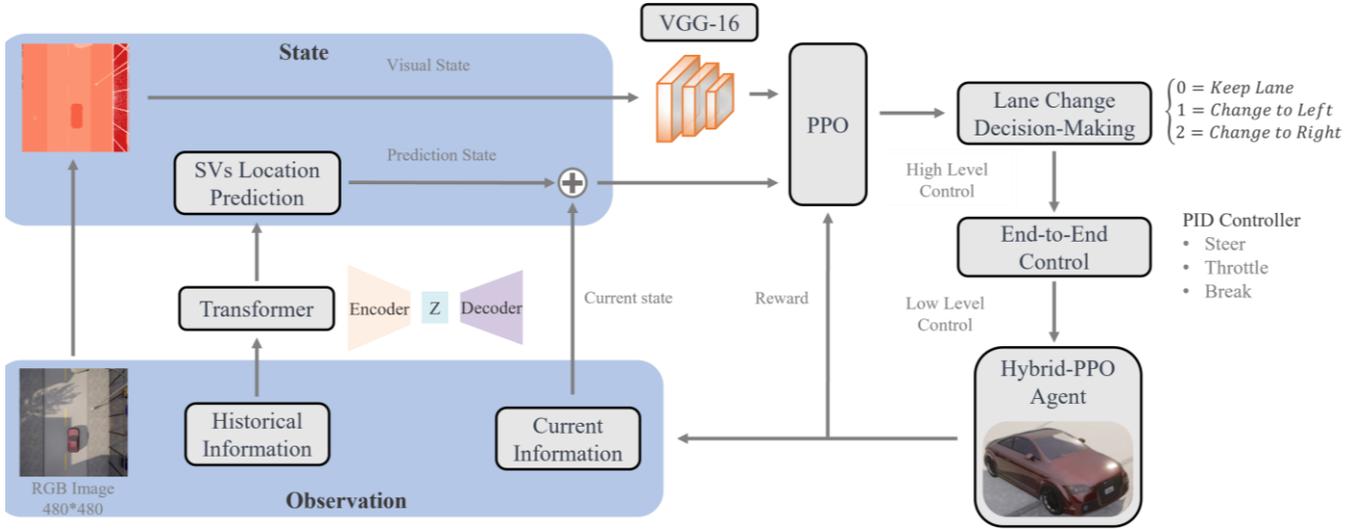

**Figure. 2:** Illustration of our approach. **Left:** The observation space and the state space. **Top middle:** Using VGG-16 to realize hybrid input. **Right:** Combination of abstract decision-making and end-to-end control to Realize the entire lane change process

$\gamma$ is discount function. At each step of MDP, a RL agent in $\gamma$ is discount function. At each step of MDP, a RL agent in state $s \in S$, takes an action $a \in A$, makes agent go to next state $s' \in S$, and get a reward r, which may be discounted by discount factor $\gamma$. The action $a$ comes from the policy $\pi(s)$, which determines the action that the agent will take in each state. The transition from state s to state $s'$ after taking action $a$ is determined by the transition probability $P$, which depends on the environment.

### B. Proximal Policy Optimization Algorithm

Reinforcement learning is a method of finding optimal policies from interactions of RL agent and the environment. The goal of reinforcement learning is maximizing the total cumulative reward of an RL agent $R_{total}$, which is usually formulated as MDP.

$$R_{total} = \sum_{t=0}^{\infty} \gamma^t r_t \quad (1)$$

**Proximal Policy Optimization (PPO)**

Proximal Policy Optimization (PPO) [22] algorithm is an on-policy gradient reinforcement learning algorithm. For policy gradient methods, the policy is parameterized in the form of $\pi(a|s;\theta)$, the policy $\pi$ is a probability distribution of action $a$ when agent at state $s$, $\theta$ is parameterized by deep neural network.

PPO using KL divergence to punish the difference between old and new policy to ensure the stability of the policy updates. While training, PPO collecting interaction experience between agent and environment to optimize the clipped surrogate objective function:

$$\hat{L}_\theta^{CLIP} = \hat{E}_t[\min(r_t(\theta)\hat{A}_t, clip(r_t(\theta, 1-\varepsilon, 1+\varepsilon)\hat{A}_t)] \quad (2)$$

where $\hat{A}_t$ is the estimation of advantage, $\hat{E}_t$ is the empirical expectation. The objective function constraints the action probability ratio in range of [1-ϵ, 1 + ϵ].

### IV. APPROACH

The goal of our work is to generate safe lane change decisions by a deep reinforcement learning agent. Even in a complex traffic flow, ego vehicle can drive smoothly through reasonable lane change. The DRL agent can make rational use of rich environmental information to generate decisions, including the current state and the potential influence of surrounding vehicles.

To show our method, we establish a complex traffic environment CARLA simulator, including a large number of slow autonomous surrounding vehicles. A deep reinforcement learning policy is executed to illustrate how the ego vehicle can smoothly move through traffic flow and get to the destination through rational lane change decisions. In the rest of this part, we will detailly explain the components of the overall model, an overview graph is shown in Fig. 2.

### A. Hybrid-PPO Algorithm

In this work, we propose a Hybrid-PPO method, which takes low-dimensional sensor data, high-dimensional image and the surrounding vehicles trajectory prediction information as state space to realize the combination of reinforcement learning method of vehicle lane changing decision.

**Transformer Trajectory Prediction**

The surrounding vehicles have an important influence on ego vehicle's lane change decision. Future trajectories of surrounding vehicles can be predicted through the historical trajectory. Therefore, we use transformer to predict trajectories of surrounding vehicles by leveraging their history locations. Transformer is a modular architecture, which has two parts, the encoder and the decoder. The sub-modules of transformer including self-attention module, feed-forward fully-connected module and residual connection module. The core self-attention mechanism formula is as follows:

$$\text{attention}(Q,K,V) = \text{softmax}(\frac{QK^T}{\sqrt{d_k}}) \quad (3)$$

where query (Q) represents one sequence and keys (K) represents all other sequences. Q is compared with K by dotting key ($d_k$) and finally output value (V). Due to its ability to capture long-range dependencies, self-attention can be effectively utilized in vehicle trajectory prediction.

In this module, the trajectories of the surrounding vehicles are input into the transformer model as a time series to finally obtain the expected location of the surrounding vehicles.

**VGG-16 for Hybrid Input**

In the CARLA simulator, we add some basic sensors on all vehicles, including speed, positioning, direction, collision sensors, and lane sensors. By collecting these data, the basic driving conditions of vehicles are present.

Besides, an extra aerial view RGB camera is attached to the ego vehicle. Therefore, in each simulation step, we can obtain both low-dimensional basic sensor data and high-dimensional RGB image. The data of the low-dimensional basic sensor are denoted by $f_{sensor} \in \mathbb{R}^{c'*l}$, where $c'$ is the information dimension, $l$ is the amount of information. In order to fuse the high-dimensional RGB image information with sensor data, the original RGB image is converted into semantic image, as shown in Fig. 2. For semantic image, we use a convolutional neural network, the VGG-16 model, to achieve feature extraction and dimension reduction.

$$x_{output}^{c'*l'} = VGG(x_{input}^{c*h*l}) \quad (4)$$

where input semantic image $x_{input}$ size is $c*h*l = 3*120*120$, and output feature $x_{output}$ size is $c'*l' = 1*10$, consistent with the size of low-dimensional sensor data.

**Hybrid-PPO**

For applying the DRL agent in this decision-making task, we define the following 1) observation space, 2) state space, 3) action space, and 4) reward function (shows in Fig. 2).

1) Observation space:

We choose a continuous observation space. At each simulation step the PPO agent obtains observation from the CARLA simulator.

**Historical observation:** Trajectory data (x-y coordinates) of last ten simulation steps of the eight nearest surrounding vehicles to the ego vehicle.

**Visual observation:** a 480*480 aerial view RGB image.

**Current observation:** the current sensor data of the ego vehicle and the eight nearest surrounding vehicles from the sensor, including vehicle direction, coordinates, speed, lane, traffic lights and other information.

2) State space:

**Prediction state:** the trajectory prediction of eight nearest surrounding vehicles to the ego vehicle after ten simulation steps. The prediction is from the transformer trajectory prediction model.

**Visual state:** a 120*120 aerial view semantic image. Semantic information is extracted from the mapping between pixel values and semantics built inside CARLA.

**Current state:** same as the current observation in observation space.

3) Action space:

We use a discrete action space to represent abstract lane change decisions, which is shown in Tab. I.

TABLE I. ACTION SPACE

| Action Space | Decision |
|---|---|
| $a_0$ | Change lane to the right |
| $a_1$ | Stay in current lane |
| $a_2$ | Change lane to the left |

4) Reward function

The reward function is designed to encourage the following behaviors: getting to the destination, avoiding traffic accidents (including collision, outside road, etc.).

The reward function is defined as follows:

$$R_{total} = R_{collide} + R_{out-road} + R_{go-forward} + R_{success} \quad (5)$$

$$\begin{cases} R_{collide} = -500 \\ R_{out-road} = -10 \\ R_{go-forward} = distance \\ R_{success} = 100/150/200/1000 \end{cases} \quad (6)$$

*B. End-to-End Control*

After the lane change decision is got from the Hybrid-PPO agent, the following end-to-end lane change action of ego vehicle is realized by a PID controller. The end-to-end lane change action of ego vehicle is realized by a PID controller based on the lane change decision made by the Hybrid-PPO agent. The start point of lane change behavior is the current location of ego vehicle, and the target point is decided by the lane change decision. A series of lane change process points are generated by trajectory planning between the start point and the target point. The PID controller receives the process points and generates low-level control instructions for the ego vehicle to attain a desired trajectory by minimizing errors. The low-level control instruction of ego vehicle including throttle, brake and direction values. Finally, the entire lane change process of ego vehicle is realized. The pseudo-code of this work is shown in Algorithm. 1

| Algorithm 1 H-PPO Training Algorithm |
|---|

1. **Initialize** the simulation environment $env$, spawn ego vehicle and surrounding vehicles.
2. **Get** initial observation $o_0$ from $env$.
3. **Predict** the position of eight nearest surrounding vehicles, **transit** RGB image into sematic image, **keep** current information, combine into state $s_0$.
4. **Initialize** the PPO policy $\pi_\theta$ with weight $\theta_0$.
5. **Instantiate** an empty replay buffer $B$ with maximum length of $l_B$.
6. **for** $i \leftarrow 0$ **to** $N$ **do**
7.     At state $s_i$ select an action (lane change decision) $a_i$ from $\pi_{\theta_{old}}(s_i)$.
8.     According to lane change decision $a$, agent do corresponding end-to-end lane change actions under PID control.
9.     Get reward $r_i$, get new observation $o_{i+1}$ and transit to new state $s_{i+1}$.
10.    Push MDP transition $[s_i, a_i, r_i, s_{i+1}]$ into the replay buffer $B$.
11.    Compute advantage estimates $\hat{A}_i, \ldots, \hat{A}_{i+7}$.
12.    **if** i % 8 == 0 **then**
13.       Optimize surrogate objective function $\hat{L}_\theta^{CLIP}$
14.       $\theta_{old} \leftarrow \theta$
15.    **if** epoch is done **then**:
16.       Record the cumulative reward.
17.    Reset $env$ and get a new initial observation $o_0$.

## V. EXPERIMENT

### A. Hardware Specs and Software

All experiments are trained on a single NVIDIA RTX3070 GPU, 12th Gen Intel(R) Core (TM) i7-12700H 2.30 GHz CPU, and 16 GB RAM.

### B. Simulation Environment

In this study, we use the CARLA [23] simulation platform to generate training and testing scenarios. We use a CARLA's built-in map *'Town06'* and randomly generate 100 black *Ford.crown* surrounding vehicles, which are all controlled by CARLA's built-in autopilot module. At the same time, a red *Audi.tt* vehicle is spawned as the ego vehicle. While driving, the ego vehicle should move forward and avoid surrounding vehicles by making safe lane changes in the traffic flow. The cycle speed of the surrounding vehicles is set at 24km/h, and the target speed of the ego vehicle is set at 30km/h. Such speed difference settings ensure that the ego vehicle will inevitably change lanes in the dense traffic flow. The simulation environment is shown in Fig. 3.

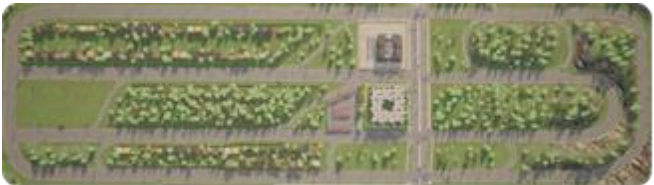

**Figure. 3:** The simulation environment in CARLA

### C. Implementation of transformer

**Trajectory dataset collection:** We collect our own trajectory dataset from CARLA simulator. We create 20 vehicles controlled by expert driver, CARLA autopilot. The data is collected as a rate of 20Hz in 2000 simulation steps.

**Evaluation metrics:** Consistent with most studies, in this paper, Mean Average Displacement (MAD, equivalent to Average Displacement Error ADE) and Final Average Displacement (FAD, equivalent to Final Displacement Error FDE) [24]. These two metrics can effectively reflect the deviation between the predicted value of the model and the ground truth.

**Prediction results:** The performance of transformer-based vehicle trajectory prediction model on validation set are shown in Tab. II.

TABLE II.     TRAJECTORY PREDICTION RESULT

| MAD | FAD |
|---|---|
| 0.382 | 0.597 |

### C. PID controller

We have defined two sets of PID parameters for lateral control and longitudinal control of the vehicle. Using CARLA built-in function *VehiclePIDController()* to transfer target points into throttle, break and wheel angel instructions. PID controller parameters are:

lateral parameters:

$$K_p = 1.95, K_p = 0.2, K_p = 0.07, \ dt = 0.1$$

longitude parameters:

$$K_p = 1, K_p = 0, K_p = 0.75, \ dt = 0.1$$

The effect of the PID controller is shown in Fig. 4, which shows the entire lane change process of a lane change decision combined with following end-to-end motion control.

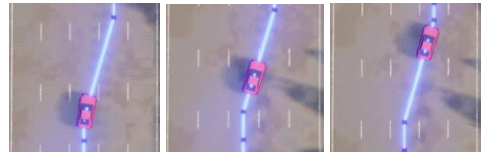

**Figure. 4:** The entire lane change process of ego vehicle

### D. Comparison Result

We compare the proposed method with two robust methods. These are the methods of Advantage Actor Critic (A2C) [14] and Deep-Q Network (DQN) [15]. Tab. III lists success rate, collision rate and time out of each method. The proposed method achieves higher score. These results confirm the reliability of the proposed method.

TABLE III.     COMPARISION EXPERIMENT RESULT

|  | Success Rate | Collision Rate | Time out |
|---|---|---|---|
| A2C [14] | 47.8% | 52.2% | 0.0% |
| DQN [15] | 46.5% | 53.3% | 0.2% |
| Hybrid-PPO | 87.8% | 10.5% | 2.7% |

## E. Ablation Study

To validate the effectiveness of our approach, ablation experiments is conducted to evaluate the contribution of different inputs. We separately test different input, including original low-dimensional current data, low-dimensional current data combined with image, low-dimensional current data with predictions, and all-hybrid input. In Fig. 5 and Fig. 6, we show the convergence curves for our approach with different state input (Hybrid-input, Prediction with low-dimensional data input, image with low-dimensional data input, only low-dimensional data input). The results in Tab. IV confirm that the state-space corresponding model, which includes picture, prediction and low-dimensional information, achieved the best performance in the lane change decision task.

TABLE IV. ABLATION EXPERIMENT RESULT

|  | Success Rate | Collision Rate | Time out |
|---|---|---|---|
| **Pure PPO** | 47.6% | 50.0% | 2.2 % |
| **Image-PPO** | 63.8% | 34.0% | 0.2% |
| **Prediction-PPO** | 76.4% | 22.7% | 0.9% |
| **Hybrid-PPO** | 87.8% | 10.5% | 2.7% |

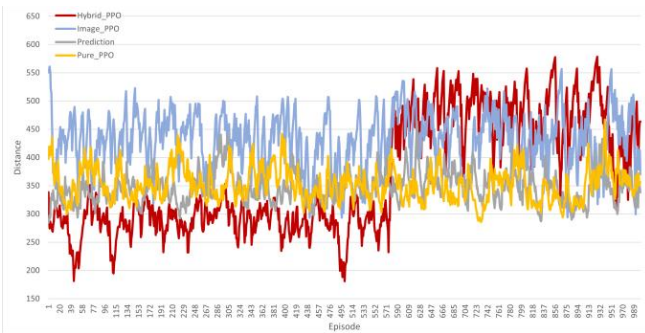

**Figure. 5:** The safe forward distance of ego vehicle

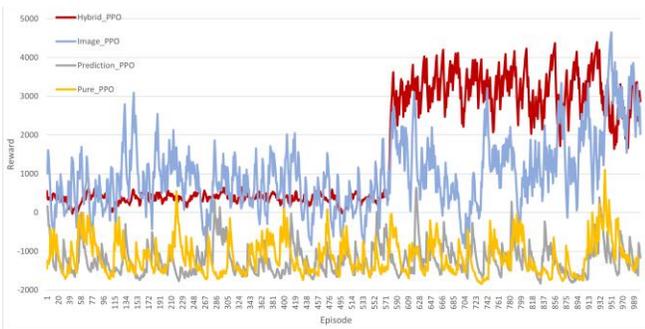

**Figure. 6:** The total reward

## VI. CONCLUSION

In summary, we proposed a deep reinforcement learning algorithm that can accept hybrid inputs to improve the safety of lane change decisions in traffic flow. In this method, the trajectory prediction of the surrounding vehicles is incorporated into the state space to reduce the danger of the potential behavior of the surrounding vehicles. Secondly, low-dimensional sensor information and high-dimensional image information are effectively fused in state space to provide more information for decision making. Finally, the combination of abstract decision making and end-to-end vehicle control ensures the integrity of the lane change action. The results of comparison experiment and ablation study proves the reliability of our method in the safe generation of lane change decisions. All these experiments were conducted in the CARLA simulator.